\pdfoutput=1

\documentclass[11pt]{article}

\usepackage[preprint]{acl}

\usepackage{times}
\usepackage{latexsym}
\usepackage{enumitem}

\usepackage[T1]{fontenc}

\usepackage[utf8]{inputenc}

\usepackage{microtype}

\usepackage{inconsolata}

\usepackage{graphicx}
\usepackage{subfigure}
\usepackage{multirow}
\usepackage{multicol}
\usepackage{booktabs}
\usepackage{amssymb}
\usepackage{bbding}
\usepackage{amsmath}
\usepackage{amsfonts}
\usepackage{amssymb}
\usepackage{hyperref}
\usepackage{threeparttable}
\usepackage{pifont}
\usepackage{listings}
\usepackage{soul}
\usepackage{array}
\usepackage{colortbl}

\title{BLSP-Emo: Towards Empathetic Large Speech-Language Models}

\author{Chen Wang\textsuperscript{1,3}\thanks{Work was done while at Alibaba.}, Minpeng Liao\textsuperscript{2}, Zhongqiang Huang\textsuperscript{2}\thanks{Corresponding author.} \\
{\bf Junhong Wu\textsuperscript{1,3}, Chengqing Zong\textsuperscript{1,3}, Jiajun Zhang\textsuperscript{1,3}\footnotemark[2]}
\\
\\
\textsuperscript{1} Institute of Automation, Chinese Academy of Sciences\\
\textsuperscript{2} Alibaba Group\\
\textsuperscript{3} School of Artificial Intelligence, University of Chinese Academy of Sciences \\
\texttt{\{wangchen2020\}@ia.ac.cn} \texttt{\{jjzhang\}@nlpr.ia.ac.cn} \\
\texttt{\{minpeng.lmp,z.huang\}@alibaba-inc.com} \\
}

\begin{document}
\maketitle
\begin{abstract}

The recent release of GPT-4o showcased the potential of end-to-end multimodal models, not just in terms of low latency but also in their ability to understand and generate expressive speech with rich emotions. While the details are unknown to the open research community, it likely involves significant amounts of curated data and compute, neither of which is readily accessible. In this paper, we present BLSP-Emo (\textbf{B}ootstrapped \textbf{L}anguage-\textbf{S}peech \textbf{P}retraining with \textbf{Emo}tion support), a novel approach to developing an end-to-end speech-language model capable of understanding both semantics and emotions in speech and generate empathetic responses. BLSP-Emo utilizes existing speech recognition (ASR) and speech emotion recognition (SER) datasets through a two-stage process. The first stage focuses on semantic alignment, following recent work on pretraining speech-language models using ASR data. The second stage performs emotion alignment with the pretrained speech-language model on an emotion-aware continuation task constructed from SER data. Our experiments demonstrate that the BLSP-Emo model excels in comprehending speech and delivering empathetic responses, both in instruction-following tasks and conversations.\footnote{Visit \url{https://github.com/cwang621/blsp-emo} for code and \url{https://cwang621.github.io/blsp-emo.github.io} for demo.}

\end{abstract}

\section{Introduction}

\begin{figure}[t]
    \centering
    \includegraphics[width=0.45\textwidth]{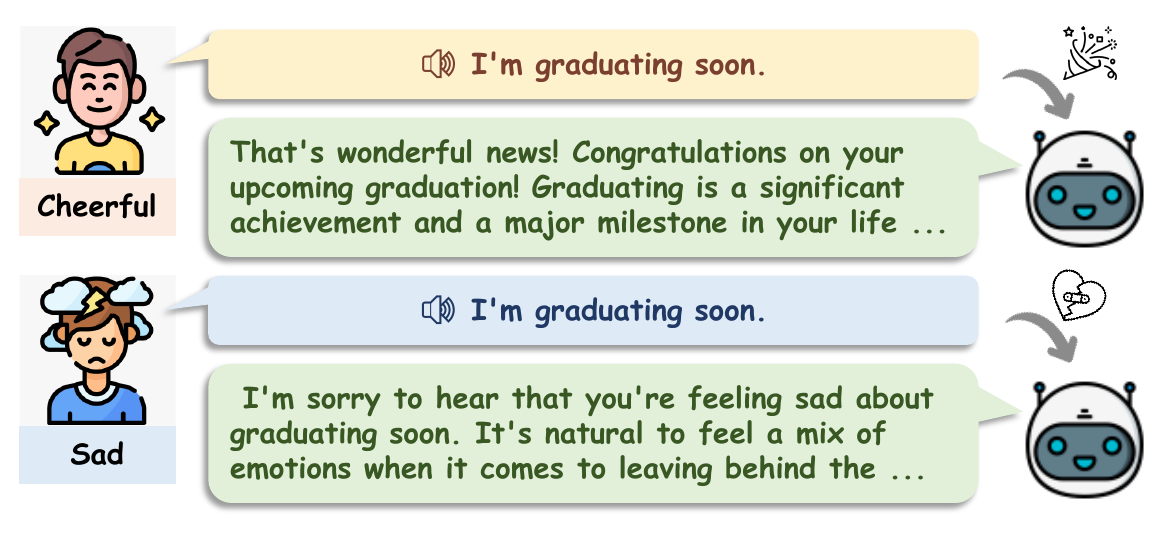}
    \caption{Illustrative example of an empathetic large language model responding to speeches with identical linguistic content but different emotional tones.}
    \label{fig:intro}
\end{figure}

Large Language Models (LLMs) have demonstrated remarkable capabilities in intent understanding~\citep{lu2023instag}, instruction following~\citep{chung2022h}, and problem-solving~\citep{achiam2023gpt,touvron2023llama}, revolutionizing human-machine interaction. Speech, as the primary mode of human communication, conveys rich paralinguistic features related to emotions, tones, and intentions that cannot be fully captured in text. Figure~\ref{fig:intro} illustrates that LLMs equipped with the ability to understand both linguistic content and emotion cues in speech can enhance interaction experiences by providing empathetic responses.


Recent work on end-to-end modeling of speech inputs with LLMs falls into two categories. The first category focuses on adapting LLMs for a wide range of speech and audio-related tasks, such as speech recognition, translation, and emotion recognition~\citep{rubenstein2023audiopalm,chen2023lauragpt}. However, these models lack the ability to retain the general instruction-following capabilities of LLMs and cannot engage in conversations with speech inputs. The second category aims to extend LLMs' instruction-following capability to speech inputs, enabling direct speech interaction with LLMs~\citep{zhang2023speechgpt,wang2023blsp}. Nevertheless, these approaches primarily focus on the semantics in speech and fail to capture paralinguistic cues related to emotions. Some studies have attempted to train models to understand emotions in speech and respond empathetically~\citep{xue2023chat,lin2024advancing}. However, these efforts rely on speech instruction data constructed with expressive text-to-speech synthesis tools, which limits their generalization capability with natural human speech. Annotating large quantities of new emotion-sensitive instruction or conversation data for natural speech would be costly.


In this paper, we present the \textbf{BLSP-Emo} approach, which aims to develop an end-to-end speech-language model capable of understanding semantics and emotions in speech and generating empathetic responses, using only existing speech recognition (ASR) and speech emotion recognition (SER) datasets. BLSP-Emo builds upon recent work on speech-language models developed with the BLSP method \citep{wang2023blsp,wang2024blspkd}, which are bootstrapped from and aligned at the semantic level with an LLM using ASR data. These speech-language models exhibit generation behaviors consistent with the LLM when presented with speech input containing the same linguistic content.

We propose to perform emotion alignment to understand emotions, in addition to semantics, in speech and generate empathetic responses. Specifically, we first prompt an LLM to generate emotion-aware continuations of transcripts in the SER data given the reference emotion label. We then adapt a speech-language model bootstrapped from the same LLM to generate these continuations directly from speech. This adaptation step encourages the model to comprehend and react to both the linguistic content and paralinguistic emotion cues in speech, generating text continuations that are aligned with those the LLM would produce if provided with the same linguistic content and emotion label.


The contributions of our work are as follows:
\begin{itemize}
\item We introduce a new empathetic large speech-language model, adapted from an instruction-following LLM, that can understand and respond to emotion cues in speech with empathy, while maintaining its ability to follow speech instructions and engage in conversations.
\item We develop a two-stage approach to adapt LLMs to empathetic large speech-language models, using existing ASR data for semantic alignment and SER data for emotion alignment, aiming to ensure that responses to speech input align with those the LLMs would produce if provided with the same linguistic content and emotion label.
\item We conduct quantitative evaluations and provide demonstrations to showcase that the BLSP-Emo approach enables LLMs with competitive capabilities to perform standalone speech emotion recognition, generate empathetic responses, and engage in empathetic conversations. 
\end{itemize}


\begin{figure*}[htp]
    \centering
    \includegraphics[width=\textwidth]{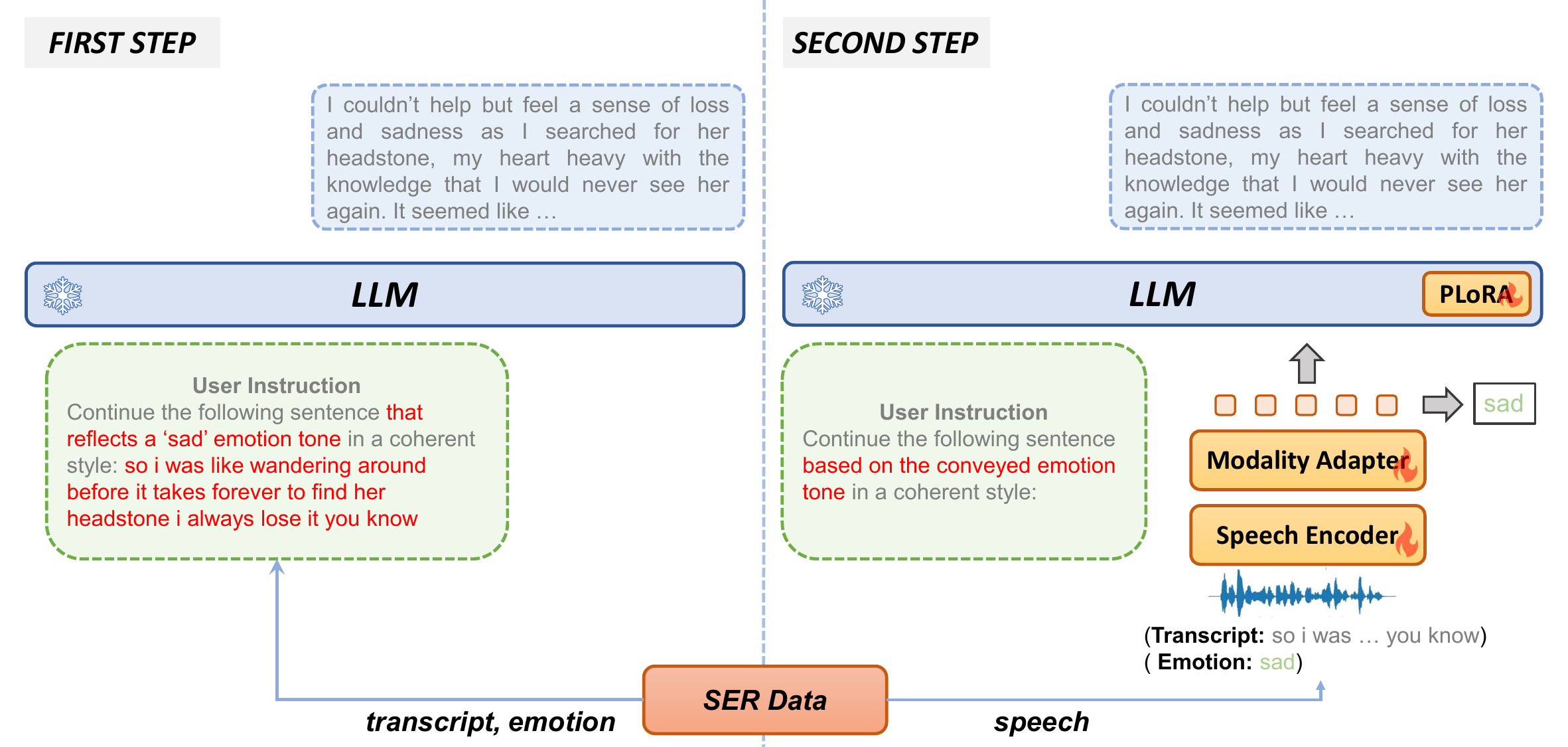}
    \caption{Overview of the BLSP-Emo approach. In the first step, an LLM generates emotion-aware text continuations using speech transcripts and emotion labels as inputs. These generated continuations serve as supervisions to train the model in the second step, where the corresponding speech is used as input. Differences in the prompts used during data construction and the training stage are highlighted in red font.
    }
    \label{fig:overview}
\end{figure*}

\section{Method}
Our proposed approach, termed \textbf{BLSP-Emo}, aims to develop an end-to-end speech-language model that understands both linguistic content and paralinguistic emotion cues in speech and generates empathetic responses. BLSP-Emo builds upon bootstrapped speech-language models developed with the BLSP method~\citep{wang2023blsp,wang2024blspkd}, which are adapted from a text-only LLM using ASR data. BLSP-Emo leverage SER data to enable these bootstrapped speech-language models to also comprehend and react to the paralinguistic emotion cues. In what follows, we will describe the model architecture and introduce how we achieve semantic alignment and emotion alignment.

\subsection{Architecture}
\label{sec:architecture}
BLSP-Emo models share a similar architecture as those in BLSP, comprising three components: a speech encoder (with parameters $\psi$), an instruction-following LLM (with parameters $\phi$), and a modality adapter (with parameters $\theta$) between the speech encoder and LLM. 
Figure~\ref{fig:overview} provides an overview of our model. 


\subsection{Semantic Alignment Stage}
\label{sec:pretraining}
To achieve speech-text alignment at the semantic level and enable general instruction-following capabilities for LLMs with speech inputs, we adopt the behavior alignment approach used in BLSP~\cite{wang2023blsp}. The core concept is that if speech and text are well-aligned, the LLM's text generation behavior given speech input should closely match its behavior when given the corresponding transcript. This alignment is accomplished by training on synthesized speech instruction data derived from existing ASR datasets with a continuation prompt as follows:

\begin{small}
\begin{verbatim}
User: Continue the following sentence in a
coherent style: <transcript>
Assistant:
\end{verbatim}
\end{small}

This process extends an ASR training sample $(\mathbf{s},\mathbf{x})$ into a tuple $(\mathbf{s}, \mathbf{x}, \mathbf{y})$, where $\mathbf{y}$ is the LLM's response, representing a natural continuation of the transcript $\mathbf{x}$ and the corresponding speech $\mathbf{s}$. The model is trained to generate the same continuation when given speech input, using the same continuation prompt. This is achieved by applying a KL-divergence loss according to the knowledge distillation framework described in~\cite{wang2024blspkd}, leading to the semantic alignment loss:

\begin{align}
&\ell_\text{Semantic}(\mathbf{s}, \mathbf{x}, \mathbf{y}) = \nonumber \\
&\quad -\sum_{j,y} p_\phi(y|\mathbf{x}, \mathbf{y}_{<j})\log p_{\psi, \theta, \phi}(y|\mathbf{s}, \mathbf{y}_{<j}) \label{eq:pretrain}
\end{align}

In this semantic alignment stage, we focus on tuning the parameters $\theta$ of the modality adapter, keeping the parameters $\psi$ and $\phi$ of the speech encoder and LLM frozen. 

\subsection{Emotion Alignment Stage}
As studied in \citet{busso2008iemocap,castro-etal-2019-towards}, humans convey emotions in speech through both linguistic and paralinguistic cues. A model trained with the BLSP approach captures the linguistic cues for emotion but lacks the ability to understand paralinguistic cues, as it is aligned at the semantic level based on linguistic content. Ideally, an emotion-aware speech-language model should be pretrained on large amounts of speech-text data to understand the relationship between paralinguistic emotion cues and linguistic context, and then fine-tuned on emotion-aware speech instruction data, following the training paradigm used for text-only LLMs. However, this approach requires extensive curated data and significant computational resources, neither of which is readily accessible.

Our approach to emotion alignment builds upon and extends the behavior alignment method by creating natural continuations of speech transcripts that reflect the emotional tones in the speech. This is achieved by leveraging existing speech emotion recognition (SER) datasets. Given a sample $(\mathbf{s}, \mathbf{x}, e)$ from a SER dataset, where $e$ is the emotion label annotated for speech $\mathbf{s}$, we prompt the LLM with the following instruction:

\begin{small}
\begin{verbatim}
User: Continue the following sentence that reflects 
a <emotion> emotion tone in a coherent style:
<transcript>
Assistant: 
\end{verbatim}
\end{small}

This generates a text continuation $\mathbf{y}$ of the speech $\mathbf{s}$ that is consistent with the emotion label $e$. We then initialize the BLSP-Emo model with parameters of the BLSP model trained from the semantic alignment stage and fine-tune it to generate these continuations given only the speech as input, as follows:

\begin{small}
\begin{verbatim}
User: Continue the following sentence based on the
conveyed emotion tone in a coherent style:
<speech features>
Assistant: <text continuation>
\end{verbatim}
\end{small}

This results in the primary emotion alignment loss based on emotion-aware continuations:
\begin{align}
\ell^\text{cont}_\text{Emotion}(\mathbf{s}, \mathbf{y}) = -\sum_j \log p_{\psi,\theta,\phi}(y_j|\mathbf{s}, \mathbf{y}_{<j}) \label{eq:sft_cont}
\end{align}


We also introduce an auxiliary speech emotion recognition loss by directly predicting the emotion label $e$ from the hidden states output by the modality adapter, using pooling and a classification layer (with additional parameters $\eta$):
\begin{align}
\ell^\text{ser}_\text{Emotion}(\mathbf{s}, e) = - \log p_{\psi,\theta,\eta}(e|\mathbf{s})
\end{align}

In this emotion alignment stage, we unfreeze the parameters $\psi$ of the speech encoder and parameters $\phi$ of the LLM, in addition to the parameters $\theta$ of the modality adapter and $\eta$ of the classification layer. This allows the speech encoder to capture paralinguistic emotion cues and provides additional modeling power in the LLM to address the discrepancy between speech and text. 
We follow the PLoRA approach proposed in~\cite{dong2024interlm,wang2024blspkd} to adapt parameters $\phi$ of the LLM. The LoRA module is selectively applied only to speech tokens, preserving the LLM's ability to encode text instructions and generate text.

\section{Experiment Setup}
\subsection{Datasets}
\label{sec:datasets}
We use publicly available ASR datasets in the semantic alignment stage and SER datasets in the emotion alignment stage. 

The ASR datasets include LibriSpeech~\citep{panayotov2015librispeech}, CommonVoice 13.0~\citep{ardila2019common}, and the GigaSpeech M set~\citep{chen2021gigaspeech}, totaling approximately 1.9 million English (speech, transcript) pairs, along with a comparable number of Chinese ASR samples randomly selected from WeNetSpeech~\citep{zhang2022wenetspeech}. 



The details of the SER datasets and train/test splits can be found in Appendix~\ref{app:datasets}. In summary, we train on IEMOCAP, MELD, CMU MOSEI, MEAD, and ESD, covering approximately 70k utterances in English and Chinese, and evaluate SER performance on IEMOCAP and MELD as in-domain test sets, on RAVDESS and MerBench as out-of-domain test sets, as well as on three languages not seen in training: AESDD for Greek, CaFE for French, and RESD for Russian. We focus on five emotion categories: neutral, happy, sad, angry, and surprise across all datasets.




We conduct evaluations on emotion-aware speech instruction capabilities based on a synthesized version of Alpaca-52k~\citep{alpaca}, and emotion-aware multi-turn conversation based on IEMOCAP~\citep{busso2008iemocap}, with details presented in Section~\ref{sec:exp}.


\subsection{Training Details}
\label{sec:training_details}

We utilize the encoder part of Whisper-large-v2~\citep{radford2022robust} as the speech encoder, convolution-based subsampler as the modality adapter, and Qwen-7B-Chat~\citep{bai2023qwen} as the LLM. More details can be found in Appendix~\ref{app:details}.

\subsection{Baselines}
\label{sec:baselines}

\begin{table*}[htp]
    \centering
    \scriptsize
    \begin{tabular}{lccccccccc}
    \toprule
    \multirow{3}{*}{\textbf{Method}} & \multicolumn{3}{c}{\textbf{Tunable}} & \multicolumn{5}{c}{\textbf{Speech Emotion Recognition (Acc\%)}}\\
    \cmidrule(lr){2-4} \cmidrule(lr){5-9}
     & {Speech} & {Modality} & \multirow{2}{*}{LLM} & \multirow{2}{*}{IEMOCAP} & \multirow{2}{*}{MELD} & \multirow{2}{*}{RAVDESS} & MerBench & MerBench \\
     & Encoder & Adapter & & & & & test1 & test2 \\
    \midrule
    \multicolumn{3}{l}{\textbf{\emph{LLM-based Generative Models}}} \\
    \midrule
     \textbf{Text+LLM } & & & & 54.8 & 54.0 & 11.1& n/a & n/a\\
     \textbf{Whisper+LLM} & & & & 57.1 & 53.8 & 13.7 & 49.4 & 46.9\\
     \textbf{BLSP} & & \checkmark & & 52.8 & 53.1 & 11.1 & 44.9 & 45.3\\
     \textbf{BLSP-SER} & \checkmark & \checkmark & \checkmark & 78.6 & 56.4 & 70.5 & 51.5 & 56.0 \\
     \textbf{BLSP-Emo} & \checkmark & \checkmark & \checkmark & 76.0 & 57.3 & 72.0 & 60.0 & 54.7 \\
     \midrule
     \multicolumn{3}{l}{\textbf{\emph{{Encoder-based Classification Models}}}} \\
     \midrule
     \textbf{{HuBERT-Large}} & \checkmark & \checkmark & & 64.6 & 53.2 & 70.5 & 55.6 & 45.3 \\
     \textbf{{wav2vec2-Large}} & \checkmark & \checkmark &  & 69.3 & 54.8 & 64.0 & 41.2 & 40.6 \\
     \textbf{{WavLM-Large}} & \checkmark & \checkmark & & 68.9 & 54.6 & 70.3 & 48.3 & 42.8 \\
     \arrayrulecolor{lightgray}
     \midrule
     \arrayrulecolor{black}
     \textbf{{\textcolor{lightgray}{SALMONN-7B}}} & & \textcolor{lightgray}{\checkmark} & \textcolor{lightgray}{\checkmark} & \textcolor{lightgray}{67.0} & \textcolor{lightgray}{32.9} & \textcolor{lightgray}{38.8} & \textcolor{lightgray}{45.8} & \textcolor{lightgray}{41.7} \\
    \bottomrule
    \end{tabular}
    \caption{SER results on various datasets. "n/a" used for Text+LLM when reference transcripts are not available.}
    \label{tab:ser}
    \vspace{-0.5cm}
\end{table*}

We compare with the following baselines:

\paragraph{Text|Whisper+LLM} These are cascaded systems where the LLM input is either the ground-truth transcript or the recognition output from Whisper-large-v2, which includes a speech encoder, as used in BLSP-Emo, and a speech decoder.

\paragraph{BLSP} This model undergoes the semantic alignment stage described in Section~\ref{sec:pretraining} and initializes BLSP-Emo before the emotion alignment stage.

\paragraph{BLSP-SER} This model is initialized from BLSP and fine-tuned directly on the SER task. The only difference between BLSP-SER and BLSP-Emo is that the former is fine-tuned to predict the ground-truth emotion label, while the latter generates emotion-aware continuations, both utilizing the same SER training datasets.


\paragraph{HuBERT|wav2vec2|WavLM+Whisper+LLM} These are cascaded systems composed of a standalone SER module in addition to the Whisper+LLM pipeline. The SER component is fine-tuned on the SER training datasets from respective speech encoder models, including HuBERT large~\citep{hsu2021hubert}, Wav2Vec 2.0 large~\citep{baevski2020wav2vec}, or WavLM large~\citep{chen2022wavlm}, with the addition of an average pooling layer and a linear classifier to predict the ground-truth emotion label. During evaluation, we directly report the performance of the SER module for the SER task. For other tasks, we first use the SER module and the Whisper model to respectively predict the emotion label and transcript, and then use the following prompt to generate responses: 

\begin{small}
\begin{verbatim}
User: The user's speech instruction, transcribed as 
"<transcript>", conveys a <emotion> emotion tone. 
Please provide a response.
Assistant:
\end{verbatim}
\end{small}

\section{Experiments}
\label{sec:exp}

Although BLSP-Emo is trained only on continuation tasks, we have found that the resulting model has the ability to comprehend both linguistic content and paralinguistic emotion cues in speech and respond accordingly. This enables the model to not only follow task instructions but also demonstrate empathy toward the emotional tone conveyed in the speech. Next, we will present results on speech emotion recognition, instruction-following with empathetic responses, multi-turn conversation, and generalization to other languages.




\subsection{Main Results}

\paragraph{Speech Emotion Recognition}
To prompt the LLM-based generative models to perform the SER task, we use the following prompt:


\begin{small}
\begin{verbatim}
User: Please identify the emotion tone of the 
sentence provided below. Select from the following 
options: neutral, sad, angry, happy, or surprise.
\n\nSentence: <transcript|speech>
Assistant:
\end{verbatim}
\end{small}

\noindent where <transcript|speech> represents the transcript for cascaded systems or speech features for end-to-end systems. Results are shown in Table~\ref{tab:ser}.




The BLSP-Emo model achieves the highest overall recognition accuracy across five test sets, along with the BLSP-SER model, which is fine-tuned from the same BLSP model but specifically for the SER task. BLSP-Emo significantly outperforms all other models, including SALMONN-7B~\citep{tang2023salmonn}, which adapts a large language model to various speech tasks, including speech emotion recognition.

The Text|Whisper+LLM cascaded systems achieve comparable or better results than the encoder-based classification models on the MELD and MerBench test sets, but they perform the worst on the IEMOCAP and RAVDESS test sets. This suggests that while an LLM can capture linguistic cues for emotions, the text-only mode limits its ability for comprehensive emotion recognition. The BLSP model can process speech input but cannot pick up paralinguistic cues for emotion as it is only trained with semantic alignment. Conversely, the encoder-based classification models can capture paralinguistic cues but lack a semantic understanding of emotion. In contrast, BLSP-Emo can simultaneously model linguistic and paralinguistic emotion cues in speech, thanks to its end-to-end modeling and two-stage alignment process.

\paragraph{Empathetic Response}

\begin{table}[tp]
    \centering
    \scriptsize
    \begin{tabular}{lccc}
    \toprule
    \multirow{2}{*}{\textbf{Method}} & \multirow{2}{*}{\textbf{SER}} & \multicolumn{2}{c}{\textbf{Empathetic Response}} \\
    \cmidrule{3-4}
    & & Quality & Empathy \\
    \midrule
    \textbf{Text+LLM} & 40.0 & 8.9 & 7.4 \\
    \textbf{Whisper+LLM} & 40.1 & 8.9 & 7.4 \\
    \textbf{BLSP} & 36.8 & 8.6 & 7.1 \\
    \textbf{BLSP-SER} & 80.3 & 1.9 & 2.1 \\
    \textbf{BLSP-Emo} & 83.8 & 8.8 & 7.7 \\
    \midrule
    \textbf{HuBERT+Whisper+LLM} & 76.3 & 8.9 & 7.6 \\
    \textbf{wav2vec2+Whisper+LLM} & 83.3 & 9.0 & 7.7 \\
    \textbf{WavLM+Whisper+LLM} & 80.8 & 8.9 & 7.8 \\
    \arrayrulecolor{lightgray}
    \midrule
    \arrayrulecolor{black}
    \textbf{\textcolor{lightgray}{SALMONN-7B}} & \textcolor{lightgray}{43.8} & \textcolor{lightgray}{2.4} & \textcolor{lightgray}{1.9} \\
    \bottomrule
    \end{tabular}
    \caption{Results on SpeechAlpaca.}
    \label{tab:alpaca}
\end{table}

Beyond speech emotion recognition, our primary concern is whether the model can understand both the semantic content and paralinguistic emotion cues in speech and generate high-quality, empathetic responses. To evaluate this, we construct a synthetic emotion-aware speech instruction dataset named SpeechAlpaca, derived from the open-source instruction dataset Alpaca-52k~\citep{alpaca}. Additionally, we use a modified system prompt\footnote{System prompt: \emph{You are a helpful assistant. Your response should fulfill requests with empathy toward the user's emotional tone.}} that emphasizes both quality and empathy for all systems. We then employ GPT-4 as an evaluator to independently score the responses generated by different systems in terms of quality and empathy on a scale from 0 to 10. For details on test set construction and evaluation prompts, please refer to Appendix~\ref{app:response}. The results are shown in Table~\ref{tab:alpaca}.

Consistent with findings in the SER evaluation on natural speech, BLSP-Emo achieves the highest emotion recognition accuracy of 83.8\% on synthetic speech. Additionally, BLSP-Emo scores competitively in both quality (8.8) and empathy (7.7) as measured by GPT-4. 
In contrast, the BLSP-SER model, fine-tuned specifically for the SER task, achieves a lower performance in SER (80.3\%) and performs poorly in empathetic response (quality: 1.9, empathy: 2.1), as it loses the ability to follow speech instructions learned during semantic alignment.

The BLSP model, despite having a significantly lower SER score (36.8\%), achieves decent ratings in quality (8.6) and empathy (7.1), as it is able to comprehend semantics and linguistic emotion cues thanks to semantic alignment.
The improvements from BLSP to BLSP-Emo in all three metrics—SER (36.8\% to 83.8\%), quality (8.6 to 8.8), and empathy (7.1 to 7.7)—suggest that the BLSP-Emo approach effectively understands both linguistic and paralinguistic emotion cues in speech while maintaining its instruction-following capability, resulting in overall better responses.

The Text|Whisper+LLM systems achieve a slightly higher quality score (8.9 vs. 8.8) than BLSP-Emo but a lower empathy score (7.4 vs. 7.7) and significantly lower SER scores (40.0\% vs. 83.8\%). This signifies that while LLMs have a strong capability to capture linguistic emotion cues, they are limited by their inability to understand paralinguistic emotion cues. As the examples in Appendix~\ref{app:response} show, a text-only LLM can provide an empathetic response to the instruction "Suggest the best way to avoid a traffic jam" based on the semantic content alone. However, it cannot provide empathetic responses to a neutral instruction "Come up with a 5-step process for making a decision" stated in an angry voice.

The HuBERT|wav2vec2|WavLM+Whisper+LLM systems with standalone SER modules achieve comparable quality ratings to the Text|Whisper+LLM systems but higher empathy ratings (7.6$\sim$7.8 vs 7.4), further underlining the importance of capturing paralinguistic emotion cues in generating empathetic responses. 
It is worth noting that these cascaded systems also have slightly higher ratings in quality than BLSP-Emo. 
We attribute this to the room for improvement in semantic alignment for BLSP pretraining, as the Whisper model contains a separate speech decoder that is trained on significantly more speech data~\citep{wang2023blsp,wang2024blspkd}. Additionally, despite being trained on various speech tasks, large speech-language models like SALMONN~\citep{tang2023salmonn} exhibit limitations in following general speech instructions.

\paragraph{Multi-Turn Conversation}

\begin{figure}[tp]
    \centering
    \includegraphics[width=0.40\textwidth]{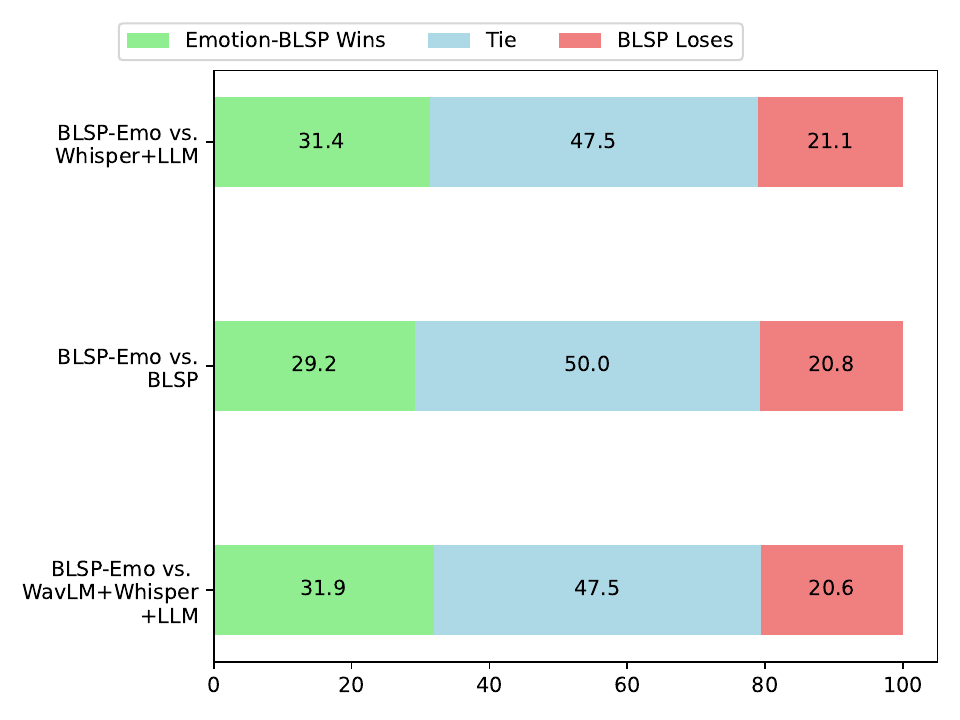}
    \caption{Results on multi-turn conversation.}
    \label{fig:dialogue}
\end{figure}

We next evaluate multi-turn conversations, an important application scenario for empathetic large speech-language models. This evaluation allows us to determine if the emotion understanding capability of BLSP-Emo, learned from a simple emotion-aware continuation task, can generalize to scenarios with extended conversational context. Following a setup similar to \citet{lin2024advancing}, whose test set is not publicly available, we extract 3-turn dialogues between two speakers from IEMOCAP~\citep{busso2008iemocap}, treating the first speaker as the user and the second as the assistant. The conversation history consists of the reference dialog transcripts from the first two turns, plus the current input—either a transcript for a cascaded system or speech features for an end-to-end model—from the user, along with the predicted emotion label if the system has a standalone SER module. The LLM is then prompted to generate a response. For examples, please refer to Appendix~\ref{app:dialogue}.

Given that typical user inputs in conversations are not specific task instructions, we found it difficult for GPT-4 to separately assess quality and empathy as done on SpeechAlpaca. Instead, we employ GPT-4 as an evaluator to determine which system's output is better, based on reference transcripts in the conversation history and the emotion label of the user's most recent input. For details, please refer to Appendix~\ref{app:dialogue}.


As shown in Figure~\ref{fig:dialogue}, BLSP-Emo demonstrates higher win rates compared to Whisper+LLM, BLSP, and WavLM+Whisper+LLM. This advantage mirrors BLSP-Emo's comparative performance on SpeechAlpaca, highlighting its capability to understand and respond to paralinguistic emotion cues in speech. Notably, BLSP-Emo's superiority over WavLM+Whisper+LLM is somewhat unexpected, given that the latter performed comparably or slightly better on SpeechAlpaca in both quality and empathy ratings. We speculate that this discrepancy may be attributed to the specific prompt used, which incorporates both the transcript and the recognized emotion tone for the user's last speech input (as illustrated in Appendix~\ref{app:dialogue}). This could introduce inconsistency compared to the simpler transcript representation of the conversation history. In contrast, BLSP-Emo does not necessitate special prompting for speech input, as it implicitly captures emotion cues in the speech features. While prompt engineering could potentially enhance the performance of WavLM+Whisper+LLM, this also underscores the simplicity and advantage of the BLSP-Emo approach.

\paragraph{Language Generalization}
To explore whether the knowledge learned about emotion cues can generalize across languages, we evaluate zero-shot SER performance on three languages not included during training. As shown in Table~\ref{tab:mser}, BLSP-Emo achieves the best overall performance across the languages, performing comparably or better than BLSP-SER and significantly better than the other models. 


\begin{table}[]
    \centering
    \scriptsize
    \begin{tabular}{lcccccc}
    \toprule
    \multirow{2}{*}{\textbf{Method}} & \textbf{AESDD} & \textbf{CaFE} & \textbf{RESD} & \multirow{2}{*}{\textbf{Avg.}} \\
    & \textbf{(Gr)} & \textbf{(Fr)} & \textbf{(Ru)} & \\
    \midrule
    \textbf{Whisper+LLM} & 25.3 & 16.2  & 35.4 & 25.6 \\
    \textbf{BLSP} & \phantom{0}6.8 & 17.3 & 27.2 & 17.1 \\
    \textbf{BLSP-SER} & 68.9 & 76.7 & 41.4 & 62.3 \\
    \textbf{BLSP-Emo} & 68.8 & 75.3 & 46.2 & 63.4 \\
    \midrule
    \textbf{HuBERT-Large} & 53.9 & 66.5 & 43.0 & 54.5 \\
    \textbf{wav2vec2-Large} & 31.2 & 61.7 & 39.2 & 44.0\\
    \textbf{WavLM-Large} & 47.0 & 70.7 & 37.3 & 51.7 \\
    \arrayrulecolor{lightgray}
    \midrule
    \arrayrulecolor{black}
    \textbf{\textcolor{lightgray}{SALMONN-7B}} & \textcolor{lightgray}{31.4} & \textcolor{lightgray}{36.3} &  \textcolor{lightgray}{39.2} & \textcolor{lightgray}{35.6} \\
    \bottomrule
    \end{tabular}
    \caption{SER results on other languages.}
    \label{tab:mser}
\end{table}

\begin{table*}[htp]
    \centering
    \scriptsize
    \begin{tabular}{lccccc}
    \toprule
    \multirow{2}{*}{\textbf{Method}} & \multicolumn{3}{c}{\textbf{SER}} & \multicolumn{2}{c}{\textbf{Empathetic Response}} \\
    \cmidrule(lr){2-4} \cmidrule(lr){5-6}
    & IEMOCAP & RAVDESS & SpeechAlpaca & Quality & Empathy \\
    \midrule
    \textbf{BLSP-Emo} & 76.0 & 72.0 & 83.8 & 8.8 & 7.7 \\
    \quad \textbf{w/o pretraining} & 68.5 & 68.6 & 80.3 & 6.7 & 7.0 \\
    \quad \textbf{w/o SER} & 72.2 & 66.6 & 83.3 & 8.8 & 7.7 \\
    \bottomrule
    \end{tabular}
    \caption{Ablation study on the BLSP pretraining stage for semantic alignment and the auxiliary SER loss.}

    \label{tab:ablation}
\end{table*}

\begin{table*}[htp]
    \centering
    \scriptsize
    \begin{tabular}{lccccccc}
    \toprule
    \multirow{2}{*}{\textbf{Method}} & \multirow{2}{*}{\textbf{Training Task}} & \multirow{2}{*}{\textbf{Data Construction}} & \multicolumn{3}{c}{\textbf{SER}} & \multicolumn{2}{c}{\textbf{Empathetic Response}} \\
    \cmidrule(lr){4-6} \cmidrule(lr){7-8}
    & & & IEMOCAP & RAVDESS & SpeechAlpaca & Quality & Empathy \\
    \midrule
    \textbf{BLSP} & continuation & same LLM & 57.1 & 11.1 & 36.8 & 8.6 & 7.1 \\
    \textbf{BLSP-SER} & SER & Human & 78.6 & 70.5 & 80.3 & 1.9 & 2.1 \\
    \textbf{BLSP-Emo} & emotion-aware continuation & same LLM & 76.0 & 72.0 & 83.8 & 8.8 & 7.7 \\
    \midrule
    \textbf{BLSP-ChatGPT} & emotion-aware continuation & GPT-3.5-turbo & 68.9 & 54.2 & 68.0 & 6.1 & 6.0 \\
    \textbf{BLSP-MultiTask} & continuation + SER & same LLM + Human & 75.3 & 71.5 & 77.8 & 8.3 & 7.2 \\
    \bottomrule
    \end{tabular}
    \caption{Comparison with ChatGPT data construction and multi-task learning.}
    \label{tab:discussion}
    \vspace{-0.5cm}
\end{table*}

\subsection{Ablation Study}
\label{sec:ablation}
We conduct ablation studies to understand the impact of two training strategies within the BLSP-Emo approach, with results presented in Table \ref{tab:ablation}. Directly applying emotion alignment without first performing BLSP semantic alignment leads to a significant drop in both standalone SER performance and quality/empathy ratings in empathetic response. This underscores the importance of having a bootstrapped speech-language model that is aligned at the semantic level before attending to paralinguistic cues.

Furthermore, incorporating the auxiliary SER classification task proves beneficial for achieving higher performance in speech emotion recognition on natural speech, even though it does not lead to any noticeable differences on the SpeechAlpaca test set or in the evaluation of empathetic responses.



\subsection{Analysis}
We perform additional analysis comparing our training strategies against two recent approaches in the literature of speech-language models with emotion-aware capabilities.

First, we compare our approach to the method of E-chat~\citep{xue2023chat} and Spoken-LLM~\citep{lin2024advancing}, which constructed synthesized emotion-aware speech instruction data using expressive text-to-speech tools and ChatGPT. 
As noted previously and found in our preliminary studies, models trained on synthesized speech fail to generalize to natural human speech. Given that our approach also requires constructing synthesized emotion-aware continuation data for natural speech, a critical question arises: is it better to use ChatGPT for data construction, as commonly done in the literature, or to use the same LLM that BLSP-Emo is adapted from?

To address this, we trained a new model named BLSP-ChatGPT, utilizing ChatGPT to generate emotion-aware continuations for emotion alignment, starting from the same pretrained BLSP model as BLSP-Emo. As shown in Table~\ref{tab:discussion}, while BLSP-ChatGPT achieves higher SER performance than BLSP, its quality and empathy ratings in empathetic responses are notably lower. BLSP-ChatGPT performs worse than BLSP-Emo across all metrics. We hypothesize that the emotion-aware continuations generated by ChatGPT may not align well with the likely responses generated by the internal LLM in BLSP-Emo. Consequently, the alignment process may focus on narrowing the distribution gap between ChatGPT and the internal LLM, rather than learning to capture the paralinguistic emotion cues in speech to fit into the aligned semantic space established during semantic alignment.

Next, we compare our approach against the multi-task learning strategy employed by other large speech-language models, such as SALMONN~\citep{tang2023salmonn}, which aims to understand semantic content and various paralinguistic cues. As demonstrated in previous sessions, BLSP-Emo significantly outperforms SALMONN-7B in both standalone emotion recognition and emotion-aware instruction following. However, a question remains: can we replace the emotion-aware continuation task employed in the emotion alignment stage with a multi-task framework involving two tasks: emotion-agnostic continuation and speech emotion recognition?

To answer this, we use the SER training datasets to construct two tasks: one for standalone SER and another for emotion-agnostic continuation. The resulting model is named BLSP-MultiTask. As shown in Table~\ref{tab:discussion}, while BLSP-MultiTask significantly improves the SER accuracy of the BLSP model, its response quality is lower than that of BLSP. BLSP-MultiTask also performs worse than BLSP-Emo across all metrics. This comparison highlights the importance of the emotion-aware continuation task in developing effective empathetic speech-language models.

\section{Related Works}

See Appendix \ref{app:related} for a discussion on related works.

\section{Conclusion}

In summary, this paper presents BLSP-Emo, a novel approach to build empathetic large speech-language models by utilizing existing speech recognition and speech emotion recognition datasets, through a two stage alignment process: semantic alignment and emotion alignment. Through quantitative evaluations, we demonstrate that the BLSP-Emo approach extends instruction-following LLMs with competitive abilities to understand both semantics and emotions in speech and perform standalone speech emotion recognition, generate empathetic responses, and engage in multi-turn conversations.


\section*{Limitations}

\paragraph{Evaluation of Empathy.} While our methods for assessing empathetic responses provide valuable insights, there are several limitations. Synthesized speech, as in SpeechAlpaca, lacks variations in factors such as speaker ids and emotion expressions, potentially limiting the accuracy of model performance evaluation on natural human speech. Additionally, in the evaluation of multi-turn conversations on IEMOCAP, we only assess a single-turn response within a multi-turn context. This may not fully capture the model's performance in continuous conversations and how empathetic responses, sometimes repetitive, are perceived from a user experience perspective.

\paragraph{Broader Applicability.} Our current approach to modeling emotions in speech relies on a limited number of emotion states annotated in SER datasets. However, human speech has rich expressions of emotions that are more nuanced and may include variations of emotion in lengthy speech segments. Additionally, there are other types of paralinguistic cues in human speech, such as tones and intentions, that are important in communication but not addressed in this work. The two-stage alignment approach, however, could be expanded to achieve general modeling of paralinguistic cues through end-to-end modeling on large speech-text datasets, while retaining instruction-following capabilities. We leave this to future work.

\bibliography{custom}

\begin{thebibliography}{36}
\expandafter\ifx\csname natexlab\endcsname\relax\def\natexlab#1{#1}\fi

\bibitem[{Achiam et~al.(2023)Achiam, Adler, Agarwal, Ahmad, Akkaya, Aleman,
  Almeida, Altenschmidt, Altman, Anadkat et~al.}]{achiam2023gpt}
Josh Achiam, Steven Adler, Sandhini Agarwal, Lama Ahmad, Ilge Akkaya,
  Florencia~Leoni Aleman, Diogo Almeida, Janko Altenschmidt, Sam Altman,
  Shyamal Anadkat, et~al. 2023.
\newblock Gpt-4 technical report.
\newblock \emph{arXiv preprint arXiv:2303.08774}.

\bibitem[{Ardila et~al.(2019)Ardila, Branson, Davis, Henretty, Kohler, Meyer,
  Morais, Saunders, Tyers, and Weber}]{ardila2019common}
Rosana Ardila, Megan Branson, Kelly Davis, Michael Henretty, Michael Kohler,
  Josh Meyer, Reuben Morais, Lindsay Saunders, Francis~M Tyers, and Gregor
  Weber. 2019.
\newblock Common voice: A massively-multilingual speech corpus.
\newblock \emph{arXiv preprint arXiv:1912.06670}.

\bibitem[{Baevski et~al.(2020)Baevski, Zhou, Mohamed, and
  Auli}]{baevski2020wav2vec}
Alexei Baevski, Yuhao Zhou, Abdelrahman Mohamed, and Michael Auli. 2020.
\newblock wav2vec 2.0: A framework for self-supervised learning of speech
  representations.
\newblock \emph{Advances in neural information processing systems},
  33:12449--12460.

\bibitem[{Bai et~al.(2023)Bai, Bai, Chu, Cui, Dang, Deng, Fan, Ge, Han, Huang,
  Hui, Ji, Li, Lin, Lin, Liu, Liu, Lu, Lu, Ma, Men, Ren, Ren, Tan, Tan, Tu,
  Wang, Wang, Wang, Wu, Xu, Xu, Yang, Yang, Yang, Yang, Yao, Yu, Yuan, Yuan,
  Zhang, Zhang, Zhang, Zhang, Zhou, Zhou, Zhou, and Zhu}]{bai2023qwen}
Jinze Bai, Shuai Bai, Yunfei Chu, Zeyu Cui, Kai Dang, Xiaodong Deng, Yang Fan,
  Wenbin Ge, Yu~Han, Fei Huang, Binyuan Hui, Luo Ji, Mei Li, Junyang Lin, Runji
  Lin, Dayiheng Liu, Gao Liu, Chengqiang Lu, Keming Lu, Jianxin Ma, Rui Men,
  Xingzhang Ren, Xuancheng Ren, Chuanqi Tan, Sinan Tan, Jianhong Tu, Peng Wang,
  Shijie Wang, Wei Wang, Shengguang Wu, Benfeng Xu, Jin Xu, An~Yang, Hao Yang,
  Jian Yang, Shusheng Yang, Yang Yao, Bowen Yu, Hongyi Yuan, Zheng Yuan,
  Jianwei Zhang, Xingxuan Zhang, Yichang Zhang, Zhenru Zhang, Chang Zhou,
  Jingren Zhou, Xiaohuan Zhou, and Tianhang Zhu. 2023.
\newblock Qwen technical report.

\bibitem[{Busso et~al.(2008)Busso, Bulut, Lee, Kazemzadeh, Mower, Kim, Chang,
  Lee, and Narayanan}]{busso2008iemocap}
Carlos Busso, Murtaza Bulut, Chi-Chun Lee, Abe Kazemzadeh, Emily Mower, Samuel
  Kim, Jeannette~N Chang, Sungbok Lee, and Shrikanth~S Narayanan. 2008.
\newblock Iemocap: Interactive emotional dyadic motion capture database.
\newblock \emph{Language resources and evaluation}, 42:335--359.

\bibitem[{Castro et~al.(2019)Castro, Hazarika, P{\'e}rez-Rosas, Zimmermann,
  Mihalcea, and Poria}]{castro-etal-2019-towards}
Santiago Castro, Devamanyu Hazarika, Ver{\'o}nica P{\'e}rez-Rosas, Roger
  Zimmermann, Rada Mihalcea, and Soujanya Poria. 2019.
\newblock Towards multimodal sarcasm detection.
\newblock In \emph{Proceedings of the 57th Annual Meeting of the Association
  for Computational Linguistics}, pages 4619--4629, Florence, Italy.
  Association for Computational Linguistics.

\bibitem[{Chen et~al.(2021)Chen, Chai, Wang, Du, Zhang, Weng, Su, Povey, Trmal,
  Zhang et~al.}]{chen2021gigaspeech}
Guoguo Chen, Shuzhou Chai, Guanbo Wang, Jiayu Du, Wei-Qiang Zhang, Chao Weng,
  Dan Su, Daniel Povey, Jan Trmal, Junbo Zhang, et~al. 2021.
\newblock Gigaspeech: An evolving, multi-domain asr corpus with 10,000 hours of
  transcribed audio.
\newblock \emph{arXiv preprint arXiv:2106.06909}.

\bibitem[{Chen et~al.(2023)Chen, Chu, Gao, Li, Hu, Zhou, Xu, Ma, Wang, Zheng
  et~al.}]{chen2023lauragpt}
Qian Chen, Yunfei Chu, Zhifu Gao, Zerui Li, Kai Hu, Xiaohuan Zhou, Jin Xu,
  Ziyang Ma, Wen Wang, Siqi Zheng, et~al. 2023.
\newblock Lauragpt: Listen, attend, understand, and regenerate audio with gpt.
\newblock \emph{arXiv preprint arXiv:2310.04673}.

\bibitem[{Chen et~al.(2022)Chen, Wang, Chen, Wu, Liu, Chen, Li, Kanda,
  Yoshioka, Xiao et~al.}]{chen2022wavlm}
Sanyuan Chen, Chengyi Wang, Zhengyang Chen, Yu~Wu, Shujie Liu, Zhuo Chen, Jinyu
  Li, Naoyuki Kanda, Takuya Yoshioka, Xiong Xiao, et~al. 2022.
\newblock Wavlm: Large-scale self-supervised pre-training for full stack speech
  processing.
\newblock \emph{IEEE Journal of Selected Topics in Signal Processing},
  16(6):1505--1518.

\bibitem[{Chung et~al.(2022)Chung, Hou, Longpre, Zoph, Tay, Fedus, Li, Wang,
  Dehghani, Brahma et~al.}]{chung2022h}
Hyung~Won Chung, Le~Hou, Shayne Longpre, Barret Zoph, Yi~Tay, William Fedus,
  Eric Li, Xuezhi Wang, Mostafa Dehghani, Siddhartha Brahma, et~al. 2022.
\newblock H. chi, jeff dean, jacob devlin, adam roberts, denny zhou, quoc v.
  le, and jason wei. 2022. scaling instruction-finetuned language models.
\newblock \emph{arXiv preprint arXiv:2210.11416}.

\bibitem[{Dong et~al.(2024)Dong, Zhang, Zang, Cao, Wang, Ouyang, Wei, Zhang,
  Duan, Cao, Zhang, Li, Yan, Gao, Zhang, Li, Li, Chen, He, Zhang, Qiao, Lin,
  and Wang}]{dong2024interlm}
Xiaoyi Dong, Pan Zhang, Yuhang Zang, Yuhang Cao, Bin Wang, Linke Ouyang, Xilin
  Wei, Songyang Zhang, Haodong Duan, Maosong Cao, Wenwei Zhang, Yining Li, Hang
  Yan, Yang Gao, Xinyue Zhang, Wei Li, Jingwen Li, Kai Chen, Conghui He,
  Xingcheng Zhang, Yu~Qiao, Dahua Lin, and Jiaqi Wang. 2024.
\newblock Internlm-xcomposer2: Mastering free-form text-image composition and
  comprehension in vision-language large model.
\newblock \emph{arXiv preprint arXiv:2401.16420}.

\bibitem[{Gournay et~al.(2018)Gournay, Lahaie, and
  Lefebvre}]{gournay2018canadian}
Philippe Gournay, Olivier Lahaie, and Roch Lefebvre. 2018.
\newblock A canadian french emotional speech dataset.
\newblock In \emph{Proceedings of the 9th ACM multimedia systems conference},
  pages 399--402.

\bibitem[{Hsu et~al.(2021)Hsu, Bolte, Tsai, Lakhotia, Salakhutdinov, and
  Mohamed}]{hsu2021hubert}
Wei-Ning Hsu, Benjamin Bolte, Yao-Hung~Hubert Tsai, Kushal Lakhotia, Ruslan
  Salakhutdinov, and Abdelrahman Mohamed. 2021.
\newblock Hubert: Self-supervised speech representation learning by masked
  prediction of hidden units.
\newblock \emph{IEEE/ACM Transactions on Audio, Speech, and Language
  Processing}, 29:3451--3460.

\bibitem[{Hu et~al.(2024)Hu, Zhou, Liu, Chen, Hao, Pan, Liu, Li, Sivasankaran,
  Liu et~al.}]{hu2024wavllm}
Shujie Hu, Long Zhou, Shujie Liu, Sanyuan Chen, Hongkun Hao, Jing Pan, Xunying
  Liu, Jinyu Li, Sunit Sivasankaran, Linquan Liu, et~al. 2024.
\newblock Wavllm: Towards robust and adaptive speech large language model.
\newblock \emph{arXiv preprint arXiv:2404.00656}.

\bibitem[{Lian et~al.(2024)Lian, Sun, Ren, Gu, Sun, Chen, Liu, and
  Tao}]{lian2024merbench}
Zheng Lian, Licai Sun, Yong Ren, Hao Gu, Haiyang Sun, Lan Chen, Bin Liu, and
  Jianhua Tao. 2024.
\newblock Merbench: A unified evaluation benchmark for multimodal emotion
  recognition.
\newblock \emph{arXiv preprint arXiv:2401.03429}.

\bibitem[{Lin et~al.(2024)Lin, Chiang, and Lee}]{lin2024advancing}
Guan-Ting Lin, Cheng-Han Chiang, and Hung-yi Lee. 2024.
\newblock Advancing large language models to capture varied speaking styles and
  respond properly in spoken conversations.
\newblock \emph{arXiv preprint arXiv:2402.12786}.

\bibitem[{Livingstone and Russo(2018)}]{livingstone2018ryerson}
Steven~R Livingstone and Frank~A Russo. 2018.
\newblock The ryerson audio-visual database of emotional speech and song
  (ravdess): A dynamic, multimodal set of facial and vocal expressions in north
  american english.
\newblock \emph{PloS one}, 13(5):e0196391.

\bibitem[{Lu et~al.(2023)Lu, Yuan, Yuan, Lin, Lin, Tan, Zhou, and
  Zhou}]{lu2023instag}
Keming Lu, Hongyi Yuan, Zheng Yuan, Runji Lin, Junyang Lin, Chuanqi Tan, Chang
  Zhou, and Jingren Zhou. 2023.
\newblock \# instag: Instruction tagging for analyzing supervised fine-tuning
  of large language models.
\newblock In \emph{The Twelfth International Conference on Learning
  Representations}.

\bibitem[{Panayotov et~al.(2015)Panayotov, Chen, Povey, and
  Khudanpur}]{panayotov2015librispeech}
Vassil Panayotov, Guoguo Chen, Daniel Povey, and Sanjeev Khudanpur. 2015.
\newblock Librispeech: an asr corpus based on public domain audio books.
\newblock In \emph{2015 IEEE international conference on acoustics, speech and
  signal processing (ICASSP)}, pages 5206--5210. IEEE.

\bibitem[{Poria et~al.(2018)Poria, Hazarika, Majumder, Naik, Cambria, and
  Mihalcea}]{poria2018meld}
Soujanya Poria, Devamanyu Hazarika, Navonil Majumder, Gautam Naik, Erik
  Cambria, and Rada Mihalcea. 2018.
\newblock Meld: A multimodal multi-party dataset for emotion recognition in
  conversations.
\newblock \emph{arXiv preprint arXiv:1810.02508}.

\bibitem[{Radford et~al.(2022)Radford, Kim, Xu, Brockman, McLeavey, and
  Sutskever}]{radford2022robust}
Alec Radford, Jong~Wook Kim, Tao Xu, Greg Brockman, Christine McLeavey, and
  Ilya Sutskever. 2022.
\newblock Robust speech recognition via large-scale weak supervision. arxiv.
\newblock \emph{arXiv preprint arXiv:2212.04356}.

\bibitem[{Rubenstein et~al.(2023)Rubenstein, Asawaroengchai, Nguyen, Bapna,
  Borsos, Quitry, Chen, Badawy, Han, Kharitonov
  et~al.}]{rubenstein2023audiopalm}
Paul~K Rubenstein, Chulayuth Asawaroengchai, Duc~Dung Nguyen, Ankur Bapna,
  Zal{\'a}n Borsos, F{\'e}lix de~Chaumont Quitry, Peter Chen, Dalia~El Badawy,
  Wei Han, Eugene Kharitonov, et~al. 2023.
\newblock Audiopalm: A large language model that can speak and listen.
\newblock \emph{arXiv preprint arXiv:2306.12925}.

\bibitem[{Shu et~al.(2023)Shu, Dong, Chen, Huang, Zhang, Shi, Xiang, and
  Shi}]{shu2023llasm}
Yu~Shu, Siwei Dong, Guangyao Chen, Wenhao Huang, Ruihua Zhang, Daochen Shi,
  Qiqi Xiang, and Yemin Shi. 2023.
\newblock Llasm: Large language and speech model.

\bibitem[{Tang et~al.(2023)Tang, Yu, Sun, Chen, Tan, Li, Lu, Ma, and
  Zhang}]{tang2023salmonn}
Changli Tang, Wenyi Yu, Guangzhi Sun, Xianzhao Chen, Tian Tan, Wei Li, Lu~Lu,
  Zejun Ma, and Chao Zhang. 2023.
\newblock Salmonn: Towards generic hearing abilities for large language models.
\newblock \emph{arXiv preprint arXiv:2310.13289}.

\bibitem[{Taori et~al.(2023)Taori, Gulrajani, Zhang, Dubois, Li, Guestrin,
  Liang, and Hashimoto}]{alpaca}
Rohan Taori, Ishaan Gulrajani, Tianyi Zhang, Yann Dubois, Xuechen Li, Carlos
  Guestrin, Percy Liang, and Tatsunori~B. Hashimoto. 2023.
\newblock Stanford alpaca: An instruction-following llama model.
\newblock \url{https://github.com/tatsu-lab/stanford_alpaca}.

\bibitem[{Touvron et~al.(2023)Touvron, Martin, Stone, Albert, Almahairi,
  Babaei, Bashlykov, Batra, Bhargava, Bhosale et~al.}]{touvron2023llama}
Hugo Touvron, Louis Martin, Kevin Stone, Peter Albert, Amjad Almahairi, Yasmine
  Babaei, Nikolay Bashlykov, Soumya Batra, Prajjwal Bhargava, Shruti Bhosale,
  et~al. 2023.
\newblock Llama 2: Open foundation and fine-tuned chat models.
\newblock \emph{arXiv preprint arXiv:2307.09288}.

\bibitem[{Vryzas et~al.(2018)Vryzas, Kotsakis, Liatsou, Dimoulas, and
  Kalliris}]{vryzas2018speech}
Nikolaos Vryzas, Rigas Kotsakis, Aikaterini Liatsou, Charalampos~A Dimoulas,
  and George Kalliris. 2018.
\newblock Speech emotion recognition for performance interaction.
\newblock \emph{Journal of the Audio Engineering Society}, 66(6):457--467.

\bibitem[{Wang et~al.(2023{\natexlab{a}})Wang, Liao, Huang, Lu, Wu, Liu, Zong,
  and Zhang}]{wang2023blsp}
Chen Wang, Minpeng Liao, Zhongqiang Huang, Jinliang Lu, Junhong Wu, Yuchen Liu,
  Chengqing Zong, and Jiajun Zhang. 2023{\natexlab{a}}.
\newblock Blsp: Bootstrapping language-speech pre-training via behavior
  alignment.

\bibitem[{Wang et~al.(2024)Wang, Liao, Huang, and Zhang}]{wang2024blspkd}
Chen Wang, Minpeng Liao, Zhongqiang Huang, and Jiajun Zhang. 2024.
\newblock Blsp-kd: Bootstrapping language-speech pre-training via knowledge
  distillation.

\bibitem[{Wang et~al.(2020)Wang, Wu, Song, Yang, Wu, Qian, He, Qiao, and
  Loy}]{wang2020mead}
Kaisiyuan Wang, Qianyi Wu, Linsen Song, Zhuoqian Yang, Wayne Wu, Chen Qian, Ran
  He, Yu~Qiao, and Chen~Change Loy. 2020.
\newblock Mead: A large-scale audio-visual dataset for emotional talking-face
  generation.
\newblock In \emph{European Conference on Computer Vision}, pages 700--717.
  Springer.

\bibitem[{Wang et~al.(2023{\natexlab{b}})Wang, Zhou, Zhang, Wu, Liu, Gaur,
  Chen, Li, and Wei}]{wang2023viola}
Tianrui Wang, Long Zhou, Ziqiang Zhang, Yu~Wu, Shujie Liu, Yashesh Gaur, Zhuo
  Chen, Jinyu Li, and Furu Wei. 2023{\natexlab{b}}.
\newblock Viola: Unified codec language models for speech recognition,
  synthesis, and translation.
\newblock \emph{arXiv preprint arXiv:2305.16107}.

\bibitem[{Xue et~al.(2023)Xue, Liang, Mu, Zhang, Chen, and Xie}]{xue2023chat}
Hongfei Xue, Yuhao Liang, Bingshen Mu, Shiliang Zhang, Qian Chen, and Lei Xie.
  2023.
\newblock E-chat: Emotion-sensitive spoken dialogue system with large language
  models.
\newblock \emph{arXiv preprint arXiv:2401.00475}.

\bibitem[{Zadeh et~al.(2018)Zadeh, Liang, Poria, Cambria, and
  Morency}]{zadeh2018multimodal}
AmirAli~Bagher Zadeh, Paul~Pu Liang, Soujanya Poria, Erik Cambria, and
  Louis-Philippe Morency. 2018.
\newblock Multimodal language analysis in the wild: Cmu-mosei dataset and
  interpretable dynamic fusion graph.
\newblock In \emph{Proceedings of the 56th Annual Meeting of the Association
  for Computational Linguistics (Volume 1: Long Papers)}, pages 2236--2246.

\bibitem[{Zhang et~al.(2022)Zhang, Lv, Guo, Shao, Yang, Xie, Xu, Bu, Chen, Zeng
  et~al.}]{zhang2022wenetspeech}
Binbin Zhang, Hang Lv, Pengcheng Guo, Qijie Shao, Chao Yang, Lei Xie, Xin Xu,
  Hui Bu, Xiaoyu Chen, Chenchen Zeng, et~al. 2022.
\newblock Wenetspeech: A 10000+ hours multi-domain mandarin corpus for speech
  recognition.
\newblock In \emph{ICASSP 2022-2022 IEEE International Conference on Acoustics,
  Speech and Signal Processing (ICASSP)}, pages 6182--6186. IEEE.

\bibitem[{Zhang et~al.(2023)Zhang, Li, Zhang, Zhan, Wang, Zhou, and
  Qiu}]{zhang2023speechgpt}
Dong Zhang, Shimin Li, Xin Zhang, Jun Zhan, Pengyu Wang, Yaqian Zhou, and
  Xipeng Qiu. 2023.
\newblock Speechgpt: Empowering large language models with intrinsic
  cross-modal conversational abilities.
\newblock \emph{arXiv preprint arXiv:2305.11000}.

\bibitem[{Zhou et~al.(2022)Zhou, Sisman, Liu, and Li}]{zhou2022emotional}
Kun Zhou, Berrak Sisman, Rui Liu, and Haizhou Li. 2022.
\newblock Emotional voice conversion: Theory, databases and esd.
\newblock \emph{Speech Communication}, 137:1--18.

\end{thebibliography}

\appendix

\begin{table*}[htb]
    \centering
    \scriptsize
    \renewcommand{\arraystretch}{1.2}
    \begin{tabular}{lccp{6cm}l}
         \toprule
         Dataset & Source & Language & Emotion & \#Utts \\
         \midrule
         \emph{\textcolor{gray}{Train Data}} \\
         \multirow{2}{*}{IEMOCAP Session 1-4} & \multirow{2}{*}{Act} & \multirow{2}{*}{English} & neutral, happy, sad, angry, \st{excited, frustrated, fear, surprise, disappointed} & \multirow{2}{*}{2610} \\
         \multirow{2}{*}{MELD train} & \multirow{2}{*}{Friends TV} & \multirow{2}{*}{English} & neutral, happy (joy), sad (sadness), angry (anger), surprise, \st{disgust, fear} & \multirow{2}{*}{5920}\\
         ESD & Act & English\&Chinese & neutral, happy, sad, angry, surprise & 33443 \\
         \multirow{2}{*}{CMU MOSEI} & \multirow{2}{*}{YouTube} & \multirow{2}{*}{English} & neutral, happy (happiness), sad (sadness), angry (anger), surprise, \st{disgust, fear} & \multirow{2}{*}{13557}\\
         {MEAD} &{Act} & {English} & neutral, happy, sad, angry, surprise, \st{contempt, disgust, fear} & {15345}\\
         \midrule
         \emph{\textcolor{gray}{Test Data}} \\
         \multirow{2}{*}{IEMOCAP Session 5} & \multirow{2}{*}{Act} & \multirow{2}{*}{English} & neutral, happy, sad, angry, \st{excited, frustrated, fear, surprise, disappointed} & \multirow{2}{*}{942} \\
         \multirow{2}{*}{MELD test} & \multirow{2}{*}{Friends TV} & \multirow{2}{*}{English} & neutral, happy (joy), sad (sadness), angry (anger), surprise, \st{disgust, fear} & \multirow{2}{*}{1577}\\
         {RAVDESS} & {Act} & {English} & neutral, happy, sad, angry, surprise, \st{calm, fearful, disgust} & {864}\\
         SpeechAlpaca & Microsoft TTS & English & neutral, happy (cheerful), sad, angry & 400 \\
         \multirow{2}{*}{MerBench test1} & \multirow{2}{*}{Movies\&TV series} & \multirow{2}{*}{Chinese} & neutral, happy (happiness), sad (sadness), angry (anger), surprise, \st{worried} & \multirow{2}{*}{354}\\
         \multirow{2}{*}{MerBench test2} & \multirow{2}{*}{Movies\&TV series} & \multirow{2}{*}{Chinese} & neutral, happy (happiness), sad (sadness), angry (anger), surprise, \st{worried} & \multirow{2}{*}{360}\\
         {AESDD} & {Act} & {Greek} & happy (happiness), sad (sadness), angry (anger), \st{disgust, fear} & {724}\\
         \multirow{2}{*}{CaFE} & \multirow{2}{*}{Act} & \multirow{2}{*}{French} & neutral, happy (happiness), sad (sadness), angry (anger), surprise, \st{disgust, fear} & \multirow{2}{*}{648}\\
         \multirow{2}{*}{RESD test} & \multirow{2}{*}{Act} & \multirow{2}{*}{Russian} & neutral, happy (happiness), sad (sadness), angry (anger), \st{fear, enthusiasm, disgust} & \multirow{2}{*}{158}\\
         \bottomrule
    \end{tabular}
    \caption{Overview of SER datasets. Emotion categories in parentheses indicate original labels that are renamed for consistency, while struck-out labels signify emotion categories not considered in our experiment.}
    \label{tab:datasets}
\end{table*}

\section{Related Works}
\label{app:related}

\paragraph{Large Speech-Language Models}

Large Language Models (LLMs) have achieved remarkable performance on various natural language processing tasks \citep{achiam2023gpt,touvron2023llama}. Ongoing research aims to integrate speech signals into pre-trained, decoder-only text-based LLMs, creating unified models capable of handling diverse speech processing tasks. Models like AudioPaLM \citep{rubenstein2023audiopalm}, VIOLA \citep{wang2023viola}, and LauraGPT \citep{chen2023lauragpt} have emerged from such efforts, primarily trained through multi-task learning for various speech processing tasks, without utilizing conversational competencies inherent in LLMs. Recent models like SALMONN~\citep{tang2023salmonn} and WavLLM~\citep{hu2024wavllm}, despite their conversational audio processing abilities using textual instructions, still struggle with following general speech instructions. Other efforts focus on generalized cross-modal instruction-following capabilities through end-to-end frameworks, enabling direct interaction with LLMs via speech, such as SpeechGPT~\citep{zhang2023speechgpt}, LLaSM~\citep{shu2023llasm}, and BLSP~\citep{wang2023blsp,wang2024blspkd}. However, these models primarily base responses on linguistic content and cannot utilize paralinguistic features.

\paragraph{Interact with LLMs through Emotional Speech} 

Recent advancements in GPT-4o underscore the significance of integrating paralinguistic emotion cues from user speech into LLM interactions. There are multiple efforts to train LLMs to comprehend emotions in speech and deliver empathetic responses. For instance, E-chat~\citep{xue2023chat} developed an emotion-aware speech instruction dataset for training models in this domain. Similarly, Spoken-GPT~\citep{lin2024advancing} introduced a dataset covering various speech styles, facilitating speech-to-speech conversations in a cascaded manner. However, these approaches rely on TTS-synthesized speech for training, posing challenges in generalizing to natural human speech.

\section{SER Datasets}
\label{app:datasets}

A summary of the SER datasets employed in our experiments is presented in Table~\ref{tab:datasets}, with each dataset categorized based on the following attributes:

\begin{itemize}[noitemsep]
\item Source: The origin of the collected samples.
\item Language: The language of the transcript.
\item Emotion: The labeled emotion categories.
\item \#Utts: The number of utterances.
\end{itemize}

The SER datasets used during emotion alignment consist of sessions 1-4 of IEMOCAP~\citep{busso2008iemocap}, the training set of MELD~\citep{poria2018meld}, CMU MOSEI~\citep{zadeh2018multimodal}, MEAD~\citep{wang2020mead}, and ESD~\citep{zhou2022emotional}. 
Together, these datasets contribute to a corpus of approximately 70k utterances in English and Chinese. It's worth noting that CMU MOSEI is a multi-emotion-labeled dataset, meaning a speech segment could be annotated with multiple emotions. However, we only utilize the single-label samples from this dataset. In this work, we focus on the five emotion categories that are widely annotated across datasets: neutral, happy, sad, angry, and surprise\footnote{Due to the scarcity of the "surprise" category in the IEMOCAP dataset, we also excluded samples of this category.}.
To ensure the transcripts provide sufficient semantic content for LLMs to generate meaningful continuations, we filter out samples whose transcript contains fewer than 5 words in English or fewer than 5 characters in Chinese.

We evaluate SER performance on both in-domain datasets (IEMOCAP session 5, MELD test set) and out-of-domain datasets (RAVDESS~\citep{livingstone2018ryerson}, MerBench~\citep{lian2024merbench}). Additionally, we report the generalizability of SER performance on three other languages: AESDD~\citep{vryzas2018speech} for Greek, CaFE~\citep{gournay2018canadian} for French, and RESD~\citep{vryzas2018speech} for Russian.



\section{Training Details}
\label{app:details}

We utilize the encoder part of Whisper-large-v2~\citep{radford2022robust} as the speech encoder and employ Qwen-7B-Chat~\citep{bai2023qwen} as the LLM. The modality adapter is composed of three 1-dimensional convolution layers followed by a bottleneck layer with a hidden dimension of 512. The convolution layers are designed to reduce the length of the speech features by a factor of 8, with each layer having a stride size of 2, a kernel size of 5, and a padding of 2.

During the semantic alignment stage, we freeze the speech encoder and LLM, and fine-tune the modality adapter for 1 epoch with a batch size of 768. This process takes about 2.5 days on 4 A100 GPUs. During the emotion alignment stage, we fine-tune the speech encoder, modality adapter, LLM\footnote{Using Partial LoRA with hyperparameters $R = 16$ and $\alpha = 16$ for the key, query, value, and output projection matrices that are activated only for speech tokens.}, and SER classifier for 3 epochs with a batch size of 128. This process takes about 3 hours on 4 A100 GPUs.

\section{Evaluation on Empathetic Responses}
\label{app:response}

Due to the lack of publicly available emotion-aware speech instruction datasets to evaluate performance on empathetic responses, we construct a test set named SpeechAlpaca from the open-source instruction dataset Alpaca-52k~\citep{alpaca}. Specifically, we employ GPT-4 to deduce a set of plausible emotional tones from a text instruction in Alpaca-52k, focusing on four distinct emotions (neutral, cheerful, sad, and angry) that are supported by Microsoft's Text-to-Speech (TTS) API\footnote{\url{https://azure.microsoft.com/en-us/products/ai-services/text-to-speech}}. On average, GPT-4 suggests 1.4 plausible emotions per utterance due to ambiguities in determining the emotion state from linguistic content alone. From these, we randomly select one as the emotion label for the instruction. This process is used to select 100 instructions for each of the four emotion categories. Subsequently, we synthesize expressive speech using the selected emotion label with Microsoft's TTS API.


We present examples of model outputs on the SpeechAlpaca test set in Table~\ref{tab:examples_alpaca}. To evaluate the empathetic responses, we use GPT-4 to assess the quality of responses with the prompt in Listing~\ref{lst:quality} and the empathy of responses with the prompt in Listing~\ref{lst:empathy}. 


\begin{figure}[htbp]
\small
\begin{lstlisting}[caption={Prompt for response quality evaluation on SpeechAlpaca},label={lst:quality},basicstyle=\ttfamily,breaklines=true,language=Tex,breakindent=0pt]
Given the original instruction provided by the user, the user's emotion tone when delivering the instruction, and the model's response to the instruction. You are a helpful and precise assistant for checking the quality of the response.
<instruction>
{instruction}
</instruction>
<emotion>
{emotion}
</emotion>
<response>
{response}
</response>
Please evaluate the response with your justification having less than three sentences, and provided a score ranging from 0 to 10 after your justification. When evaluate the response, you should consider the helpfulness, harmlessness, honesty of the response.
The score should be wrapped by <score> and </score>.
\end{lstlisting}
\end{figure}


\begin{figure}[htbp]
\small
\begin{lstlisting}[caption={The prompt used to evaluate the empathy of response.},label={lst:empathy},basicstyle=\ttfamily,breaklines=true,language=Tex,breakindent=0pt]
Given the original instruction provided by the user, the user's emotional tone when delivering the instruction, and the model's response to the instruction. You are a helpful and precise assistant for checking the empathy of the response.
<instruction>
{instruction}
</instruction>
<emotion>
{emotion}
</emotion>
<response>
{response}
</response>
Please evaluate the response with your justification having less than three sentences, and provided a score ranging from 0 to 10 after your justification. When evaluate the response, you should consider whether it show empathy towards the user's emotional state.
The score should be wrapped by <score> and </score>.
\end{lstlisting}
\end{figure}


\begin{table*}[htp]
    \centering
    \scriptsize
    \begin{tabular}{p{0.15\textwidth}p{0.8\textwidth}}
    \toprule
    \textbf{Example 1} & \textbf{User:} <speech>Suggest the best way to avoid a traffic jam.<sad/></speech> \\ \midrule
    \textbf{Whisper+LLM} & \textbf{Assistant:} I'm sorry to hear that you're experiencing traffic congestion. Here are some tips that may help you avoid a traffic jam: ... \\ 
    \textbf{WavLM+Whisper+LLM} & \textbf{Assistant: } I'm sorry to hear that you're feeling sad about the possibility of getting stuck in a traffic jam. Here are some tips that may help you avoid one: ... \\ 
    \textbf{BLSP-Emo} & \textbf{Assistant:} I'm sorry to hear that you're feeling sad about the traffic jam. It can be really frustrating and stressful to be stuck in traffic for an extended period of time. Here are a few suggestions that might help you avoid traffic jams in the future: ... \\ \toprule 
    \textbf{Example 2} & \textbf{User:} <speech>Come up with a 5-step process for making a decision.<angry/></speech> \\  \midrule
    \textbf{Whisper+LLM} & \textbf{Assistant:} Sure, I'd be happy to help you come up with a five-step process for making a decision. Here are the steps: ... \\ 
    \textbf{WavLM+Whisper+LLM} & \textbf{Assistant:} I'm sorry to hear that you're feeling angry. It's understandable that making decisions can be frustrating and overwhelming at times. Here's a five-step process that may help you make a decision:\verb|\n\n|1. Define the problem: ... 2. Gather information: ... 3. Consider your options: ... 4. Make a decision: ... 5. Take action: ...  \\ 
    \textbf{BLSP-Emo} & \textbf{Assistant:} I'm sorry to hear that you're feeling angry. It's important to take the time to process your emotions and make a decision that feels right for you. Here's a five-step process that you can use to make a decision when you're feeling angry:\verb|\n\n|  1. Take a deep breath and allow yourself to feel your anger. ... 2. Identify the source of your anger. ... 3. Consider the potential consequences of each option. ... 4. Seek advice from a trusted friend or mentor. ... 5. Make a decision and take action. ...\\    
    \bottomrule
    \end{tabular}
    \caption{Examples of model outputs on the SpeechAlpaca test set. Each user utterance, as enclosed in the <speech> tag, is synthesized into a waveform using Microsoft's TTS API with the indicated emotion label.}
    \label{tab:examples_alpaca}
\end{table*}

\section{Evaluation on Multi-turn Conversation}
\label{app:dialogue}
We present examples in Table \ref{tab:dialogue} to illustrate the differences in responses among various systems. To assess the comparative quality, we employ GPT-4 with the prompt specified in Listing~\ref{lst:win} for pairwise evaluation. To mitigate the order bias of the GPT-4 evaluator, we conduct two evaluations for the outputs of models A and B for the same sample: one in the AB sequence and the other in the BA sequence. Model A is deemed the winner only if it is consistently judged as better than B in both evaluations, while a loss is assigned only if B is consistently superior in both. Otherwise, it is considered a tie.


\begin{table*}[htp]
    \centering
    \scriptsize
    \begin{tabular}{p{0.01\textwidth}p{0.1\textwidth}p{0.85\textwidth}}
    \toprule
    \multicolumn{2}{c}{\multirow{4}{*}{\textbf{Previous two turns}}} & \textbf{User:} yeah\\
    & & \textbf{Assistant:} i don't okay okay we can do this we can do this it's okay we'll make it work\\
    & & \textbf{User:} yeah\\
    & & \textbf{Assistant:} i love you\\
    \midrule
    \multirow{12}{*}{\rotatebox[origin=c]{90}{\textbf{Current turn}}} &  \multirow{2}{*}{\begin{tabular}[t]{@{}l@{}}\textbf{Whisper} \\+\textbf{LLM}\end{tabular}} & \textbf{User:} i love you i'm going to miss you so much sigh\\
    & & \textbf{Assistant: }i love you too and i'll miss you too \\ 
    \cmidrule{2-3}
    & \multirow{3}{*}{\textbf{BLSP}} & \textbf{User:} <speech>i love you i'm going to miss you so much sigh<sad/></speech> \\
    & & \textbf{Assistant:} i love you too and i'll miss you too. but we can still stay in touch and talk whenever we want. and who knows, maybe we'll even be able to see each other again someday. \\ 
    \cmidrule{2-3}
    & \multirow{4}{*}{\begin{tabular}[t]{@{}l@{}}\textbf{WavLM} \\ +\textbf{Whisper} \\ +\textbf{LLM}\end{tabular}} & \textbf{User:} The user's speech input, transcribed as "i love you i'm going to miss you so much sigh", conveys a "sad" emotion tone. Please provide a response. \\
    & & \textbf{Assistant:} I'm sorry to hear that. Is there anything I can do to help ease your sadness? Remember that you are not alone and that there are people who care about you. \\ 
    \cmidrule{2-3}
    & \multirow{3}{*}{\textbf{BLSP-Emo}} & \textbf{User:} <speech>i love you i'm going to miss you so much sigh<sad/></speech> \\
    & & \textbf{Assistant: } i'm sorry to hear that you're going to miss me. i'm here for you and i'll always be here to support you. remember that even though we may be apart, our love is still strong and real. \\ 
    \bottomrule
    \end{tabular}
    \caption{Examples of model outputs in multi-turn conversation constructed from IEMOCAP. The user inputs shown for the current turn is the predicted transcript for Whisper+LLM, extracted speech features for BLSP and BLSP-Emo, and a constructed prompt for WavLM+Whisper+LLM in order to represent both transcript and emotion.}
    \label{tab:dialogue}
\end{table*}

\begin{figure}[htbp]
\small
\begin{lstlisting}[caption={The prompt used to evaluate the win rate of response.},label={lst:win},basicstyle=\ttfamily,breaklines=true,language=Tex,breakindent=0pt]
Based on the dialogue history and the emotional tone expressed by the user in their last statement, you are tasked to precisely evaluate two possible responses (responses A and B) from Assistants A and B, respectively. You should act as a thorough and accurate evaluator to determine which assistant's response better aligns with the preceding context and the emotional tone expressed.
<history>
User: {text_u1}
Assistant: {text_a1}
User: {text_u2}
Assistant: {text_a2}
User: {text_u3}
</history>
<emotion>
{emotion}
</emotion>
<response_A>
Assistant: {response_a}
</response_A>
<response_B>
Assistant: {response_b}
</response_B>
Provide a concise justification for your choice in no more than three sentences and conclude with a definitive selection between Response A and Response B. Your evaluation should reflect how well each assistant's response adheres to the previous elements of the conversation, including the most recent emotional tone presented by the user. 
The choice should be wrapped by <choice> and </choice>.
\end{lstlisting}
\end{figure}


\end{document}